\documentclass{article}

\usepackage{arxiv}
\usepackage{authblk}

\usepackage{amssymb}
\usepackage{framed,multirow}

\usepackage{latexsym}

\usepackage[utf8]{inputenc} 
\usepackage[T1]{fontenc}    
\usepackage{url}            
\usepackage{booktabs}       
\usepackage{amsfonts}       
\usepackage{microtype}      
\usepackage{xcolor}         
\usepackage{subcaption}
\usepackage{colortbl}
\usepackage{multirow}
\usepackage{tcolorbox}
\usepackage{tabularx,makecell}

\makeatletter
\def\blfootnote{\xdef\@thefnmark{}\@footnotetext}
\makeatother

\usepackage{lineno}

 \author[1,4]{Alceu Bissoto}
 \author[2]{Catarina Barata}
 \author[3,4]{Eduardo Valle}
 \author[1,4]{Sandra Avila}
 \affil[1]{Institute of Computing, University of Campinas, Brazil}

\affil[2]{Institute for Systems and Robotics, Instituto Superior Técnico, Portugal}

 \affil[3]{School of Electrical and Computing Engineering, University of Campinas, Brazil}
 
\affil[4]{Recod.ai Lab, University of Campinas, Brazil}

\title{Even Small Correlation and Diversity Shifts Pose Dataset-Bias Issues}

\begin{document}

\blfootnote{e-mails: alceubissoto@ic.unicamp.br (Alceu Bissoto),
ana.c.fidalgo.barata@tecnico.ulisboa.pt (Catarina Barata),
dovalle@dca.fee.unicamp.br (Eduardo Valle), sandra@ic.unicamp.br
(Sandra Avila)}

\maketitle


\begin{abstract}
Distribution shifts are common in real-world datasets and can affect the performance and reliability of deep learning models.
In this paper, we study two types of distribution shifts: diversity shifts, which occur when test samples exhibit patterns unseen during training, and correlation shifts, which occur when test data present a different correlation between seen invariant and spurious features.
We propose an integrated protocol to analyze both types of shifts using datasets where they co-exist in a controllable manner. Finally, we apply our approach to a real-world classification problem of skin cancer analysis, using out-of-distribution datasets and specialized bias annotations. 
Our protocol reveals three findings: 
1) Models learn and propagate correlation shifts even with low-bias training; this poses a risk of accumulating and combining unaccountable weak biases; 
2) Models learn robust features in high- and low-bias scenarios but use spurious ones if test samples have them; this suggests that spurious correlations do not impair the learning of robust features;
3) Diversity shift can reduce the reliance on spurious correlations; this is counter-intuitive since we expect biased models to depend more on biases when invariant features are missing. Our work has implications for distribution shift research and practice, providing new insights into how models learn and rely on spurious correlations under different types~of shifts.
\end{abstract}








\section{Introduction}

Diversity and correlation shifts are distribution shifts commonly present in deep learning datasets. 
The former occurs when test samples exhibit previously unseen patterns and the latter when test data present a different correlation between seen invariant and spurious features. 
These shifts co-exist in real-world datasets but are studied separately, causing current solutions to be effective for a single type of shift (or correlation, or diversity)~\cite{ye2021ood}. Additionally, correlation shifts are often studied using toy datasets to allow controlling train and test spurious correlations, causing its solutions to be less applicable to real-world problems compared to diversity shifts.

Despite being often studied separately, the joint influence of diversity and correlation shifts on datasets and solutions should not be overlooked. For example, consider the medical scenario of skin lesion analysis.
Most skin lesion data publicly available come from medical centers in USA and Australia, and expanding such collection to other regions to reduce distribution shifts is expensive and often unfeasible. Thus, for this solution to reach patients in geographical or economically disadvantaged areas, distribution shifts must be accounted for in machine learning solutions. We know that different procedures for image acquisition and the presence of clinical artifacts can introduce correlation shifts \cite{bissoto2019constructing}. Moreover, differences in population characteristics, such as skin color, also highly affect performances due to diversity shift \cite{combalia2019bcn20000}. Combined, different shifts are sufficient to cause models to fail catastrophically \cite{gomolin2020artificial}, one of the main obstacles to deploying medical solutions.

To avoid these problems and go towards inclusive and integrated distribution shift analysis, we must 1) use datasets where both correlation and diversity shifts co-exist in a controllable manner, 2) study real-world classification problems, and 3) implement an evaluation protocol appropriate for both shifts. 
To validate our approach, we extend our study to a skin cancer analysis scenario. Using out-of-distribution datasets to represent diversity shifts, we control correlation shift with artificial colored squares. For a scenario with real correlation and diversity shifts, we utilize specialized bias annotations~\cite{bissoto2020debiasing} and publicly available, peer-reviewed experimental results~\cite{bissoto2022}, confirming that our setup is representative of real-world problems.  

Our main findings are: 1)~Correlation shifts are learned and propagated to the predictions even when training presents low levels of bias, 2)~Surprisingly, diversity shift can attenuate the reliance on spurious correlations, and 3) Models fully learn robust features even in high-bias scenarios, but rely on spurious ones if test samples display the spurious feature.
Our findings have important implications for distribution shift research. They show that models can capture and rely on subtle correlations that are hard to notice or avoid in the training data. This makes current methods to remove or reduce correlation shifts ineffective or infeasible, because they require human annotations of the sources of bias, which are impossible to provide for all the subtle correlations in data.

Our contributions can be summarized as follows:

\begin{enumerate}
   \item We analyze correlation and diversity shifts in an integrated manner, evaluating different intensities of each type~of~shift.
   \item Our findings show that weak spurious correlations have a significant effect on models, but that effect can be minimized if the spurious feature changes (or is interfered~upon) during test.
   \item We verify our findings in a real-world bias case, going beyond synthetic sets.
   \item 
   We provide directions for future distribution shifts~research compatible with real-world challenges.
\end{enumerate}

\section{Related Work}
To provide the reader with a better understanding of the distribution shift literature, we start by detailing its main datasets and highlighting the differences between datasets commonly used for correlation and diversity shifts. Next, we explain the common evaluation protocol for correlation and diversity shifts, which allows the reader to understand the challenges of a joint evaluation of correlation and diversity shifts.

\subsection{Data}~\label{sec:related_data}

According to Ye et al.~\cite{ye2021ood}, the datasets studied in the literature are dominated by one kind of shift (either correlation or diversity). 
Diversity shifts datasets have domain annotated samples. Test samples display previously unseen patterns that make classification challenging, even though the label of a sample is still coherent. Most available datasets describe this type of shift. PACS~\cite{li2017deeper} contains objects of 4 domains: Photos, Art, Cartoon, and Sketches; OfficeHome~\cite{venkateswara2017deep} contains Art, ClipArt, Product, and Real; Terra Incognita~\cite{beery2020iwildcam} contains wild animals in different camera locations; PatchCamelyon17-WILDS~\cite{bandi2018detection, koh2020wilds} contains histopathology samples from $5$ different medical centers; FMoW-WILDS and PovertyMap-WILDS \cite{koh2020wilds} contain satellite imagery data with samples being collected in different continents.


Correlation shift datasets often contain attribute annotations that are potential sources of spurious correlations, or divisions into partitions (or environments) where spurious correlations' intensity varies. Table \ref{table:literaturebias} shows the most common datasets, and highlights the studied training and test biases. Generally, training biases are severe, and test biases vary, leading to no standard evaluation protocol.
Synthetic datasets manipulate the confounders in the images to control train and test spurious correlations: ColorMNIST~\cite{arjovsky2019invariant} controls the colors of the digits, and Coco-on-Places~\cite{ahmed2020systematic} exploits segmentation masks on Coco to combine with backgrounds from Places to create and control spurious correlations. 
In real-world datasets, attribute annotations can be exploited to generate biased problems. For example, in CelebA~\cite{liu2015celeba}, a classical classification problem employs the ``Male'' attribute as the target, while using ``BlondHair'' as a confounder. Due to the limited number of blond male examples, the resulting high bias can misrepresent the classification task, as both ``Male'' and ``BlondHair'' exhibit nearly perfect predictive ability. A more effective approach is demonstrated in skin lesion ``trap sets''~\cite{bissoto2020debiasing}, where authors annotate skin lesion datasets with respect to artifacts and employ an optimization procedure to construct progressively increasing biased sets.

\begin{table}[h]
\caption{Datasets used in the correlation shift literature. Training and test bias show the percentage of samples from the majority confounder group that share the same target label. Adopted training biases are very high, above 80\%, and there is no standard evaluation protocol, with varying test biases.}
\centering
\footnotesize
\begin{tabular}{lcccc}
\toprule
         Dataset  & \makecell[c]{Training \\ bias} & \makecell[c]{Target  \\ label}     & \makecell[c]{Confounder} & \makecell[c]{Test \\ bias} \\
\midrule
ColoredMNIST~\cite{arjovsky2019invariant}     & 80, 90        & Digits           & Color          & ~90 \\
Waterbirds~\cite{sagawa2019distributionally}  & 95            & Bird species     & Background     & ~50 \\
CelebA~\cite{liu2015celeba,ye2021ood}          & 95            & Blonde Hair      & Gender         & ~50        \\
NICO~\cite{he2020towards,ye2021ood}   & 96  & \makecell[c]{Animals \&\\vehicles} & Background     & ~90 \\
Coco-on-Places~\cite{ahmed2020systematic} & 80            & \makecell[c]{Animals \&\\vehicles} & Background     & 100 \\
\bottomrule
\end{tabular}
\label{table:literaturebias}
\end{table}

\subsection{Bias Evaluation and Analysis}

As previously shown, distribution shift data usually contain domain or confounders annotations. Most solutions exploit them to create environments.
Environments are groups within data that share most characteristics but differ ideally in single or few aspects \cite{arjovsky2019invariant}. There are typically two approaches for defining these environments: 1) By associating each domain~with a separate environment; or
2) By varying the correlations between confounder and target labels across environments.
To illustrate this, let's consider a medical dataset containing images, target diagnoses, and the source medical centers. Using the first approach, we could assign each medical center to a unique environment. In contrast, the second approach would require each environment to have a varying proportion of samples from each medical center, while still ensuring that images from the same centers are present across multiple environments. When dealing with diversity and correlation shifts, the first and second approaches are generally employed, respectively.

For evaluation purposes, one environment is typically excluded from training and reserved for testing. 
More specifically, for diversity shift, authors often report models' performance when leaving each of the domains out of training. In other cases, the most challenging domain is selected for test, as in PatchCamelyon-WILDS, where the test center was selected as the visually most distinct one \cite{koh2020wilds}. 
 For correlation shift, since test bias can be controlled, there is no standard procedure for defining the characteristics of test (see column ``Test bias'' in Table \ref{table:literaturebias}). Recently, a systematic evaluation approach has become the new standard.
Ahmed et al.~\cite{ahmed2020systematic} study the traditional problem where the image background color is the confounder attribute, having each color correlated to a specific target label (e.g., ColorMNIST \cite{arjovsky2019invariant}). In their paper, they evaluate the performance in different test sets: \romannumeral 1) keeping the coloring scheme, maintaining same colors correlated to same labels; \romannumeral 2) randomly coloring backgrounds with same colors; and \romannumeral 3) randomly coloring backgrounds with unseen colors; \romannumeral 1) and \romannumeral 2) allows to evaluate the model's reliance on the color information,
and \romannumeral 3) allows to verify if the presence of color bias on training caused the model to be less robust to other unknown shifts.

\section{Methodology}
Measuring the effect of debiasing solutions is challenging. 
Naively evaluating models on untreated test sets may assess the models' ability to learn spurious features instead of invariant ones. To compose the problem, datasets that allow controlling the levels of correlation and diversity shifts without oversimplifying the classification task are rare~(see Section \ref{sec:related_data}).
To study the effects of diversity and correlation shifts in deep learning datasets and models in increasingly levels of complexity of correlation and diversity shifts, we propose to study three different~cases:

\begin{enumerate}
    \item Synthetic correlation and diversity shifts.
    \item Synthetic correlation shift and real diversity shift.
    \item Real correlation and diversity shifts.
\end{enumerate}

In these experiments, we give more focus to synthetic correlation shifts (instead of synthetic diversity shifts) due to the difficulty of building datasets with varying levels of correlation shift without resorting to synthetic biases. Such challenge is also thoroughly explored in our third case, where both correlation and diversity shifts are real.
Next, we describe the modifications necessary to introduce synthetic correlation and diversity shifts, and further detail each of our three case studies.

\begin{figure*}[t]
    \centering
    \hspace{2.53cm}
    \begin{minipage}{0.01\textwidth}
    \vspace{0.7cm}
    \raggedleft
    \rotatebox[origin=l]{90}{\scriptsize{~~~~~~~~~~~Melanoma~~~~~~Benign~~~~~~~~~~Melanoma~~~~~~Benign~~~~~~~~~~~~Class A~~~~~~~~Class B}} 
    \end{minipage}%
    \hspace{-2.7cm}
    \begin{minipage}{0.99\textwidth}
    \centering
    \begin{subfigure}{0.84\textwidth}
    
    \centering
            \textbf{\footnotesize{Synthetic correlation and diversity shifts}}\\
            
            \begin{subfigure}{0.08\textwidth}
            \includegraphics[width=\linewidth]{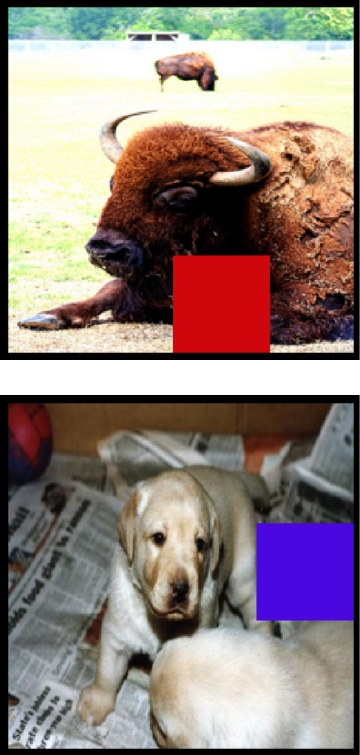}
            \end{subfigure}
            \begin{subfigure}{0.08\textwidth}
            \includegraphics[width=\linewidth]{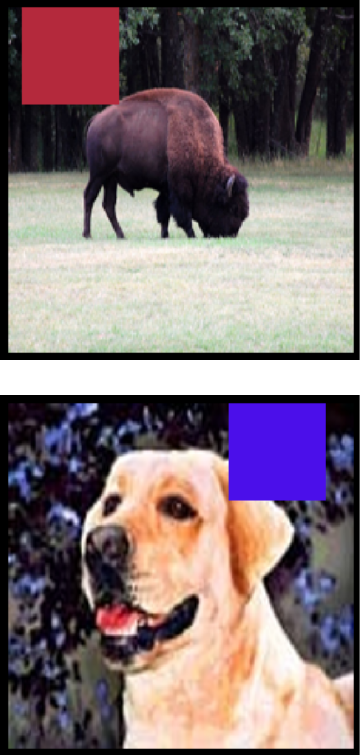}
            \end{subfigure}
            \begin{subfigure}{0.08\textwidth}
            \includegraphics[width=\linewidth]{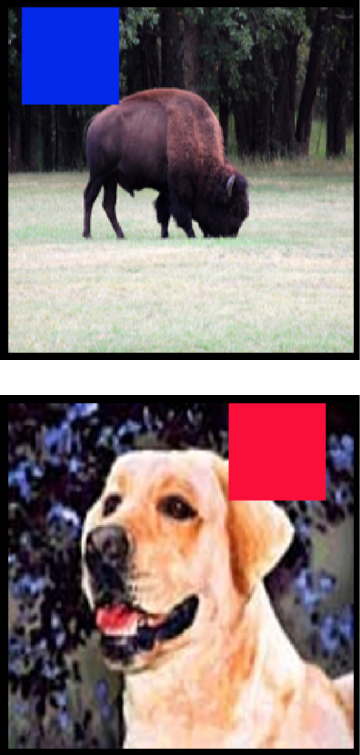}
            \end{subfigure}
            \begin{subfigure}{0.08\textwidth}
            \includegraphics[width=\linewidth]{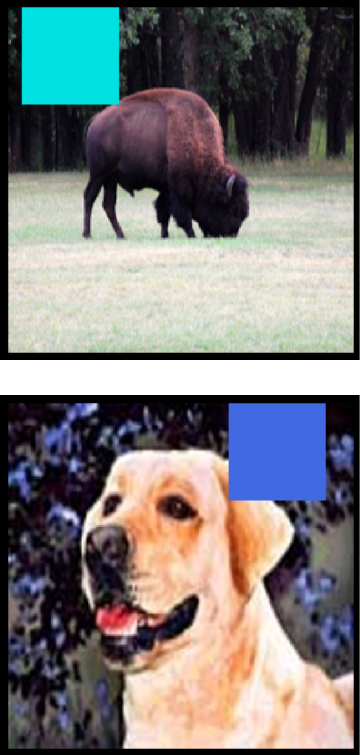}
            \end{subfigure}
            \begin{subfigure}{0.08\textwidth}
            \includegraphics[width=\linewidth]{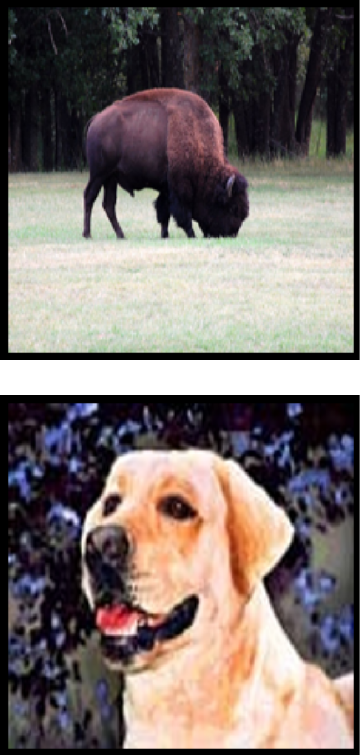}
            \end{subfigure}
            \begin{subfigure}{0.08\textwidth}
            \includegraphics[width=\linewidth]{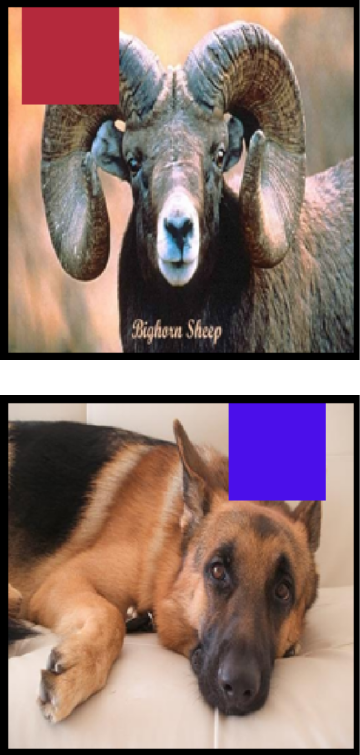}
            \end{subfigure}
            \begin{subfigure}{0.08\textwidth}
            \includegraphics[width=\linewidth]{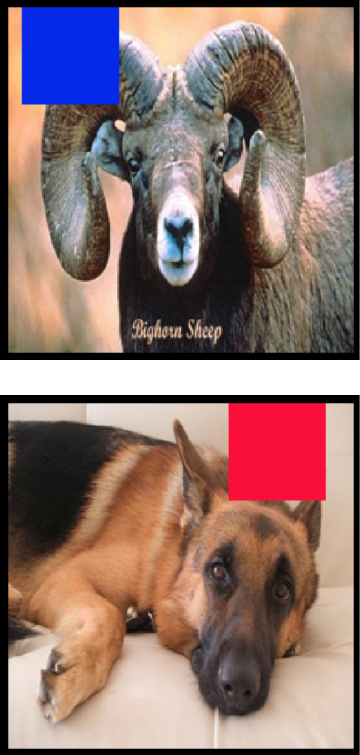}
            \end{subfigure}
            \begin{subfigure}{0.08\textwidth}
            \includegraphics[width=\linewidth]{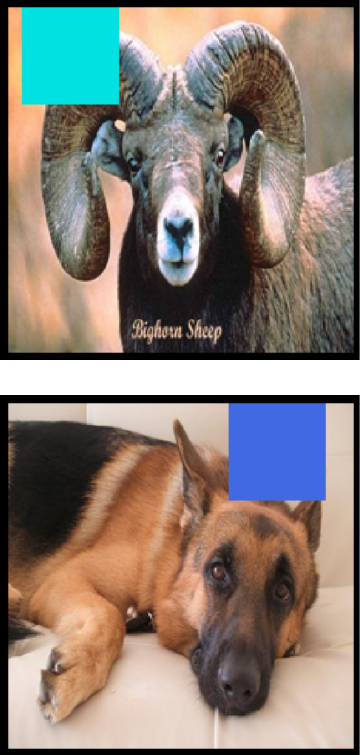}
            \end{subfigure}
            \begin{subfigure}{0.08\textwidth}
            \includegraphics[width=\linewidth]{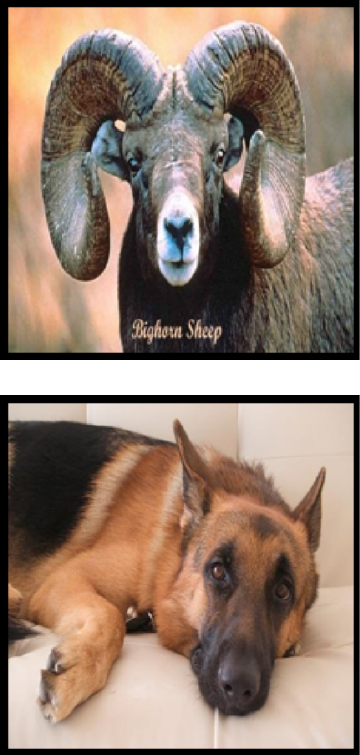}
            \end{subfigure}
            
            \vspace{0.1cm}
            \textbf{\footnotesize{Synthetic correlation, and real diversity shifts}}\\
            
            \begin{subfigure}{0.08\textwidth}
            \includegraphics[width=\linewidth]{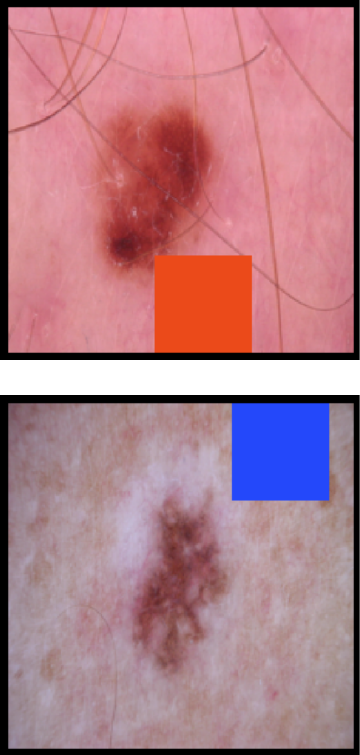}
            \end{subfigure}
            \begin{subfigure}{0.08\textwidth}
            \includegraphics[width=\linewidth]{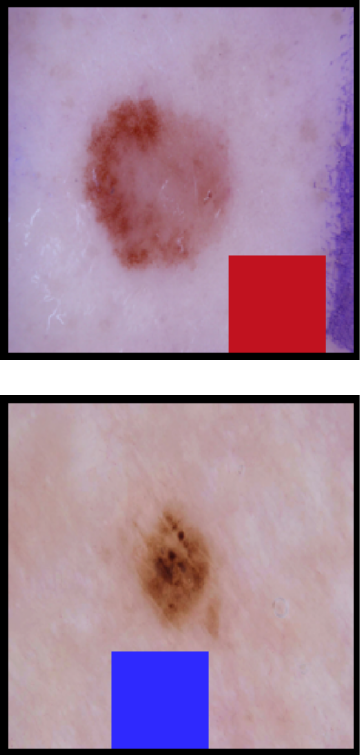}
            \end{subfigure}
            \begin{subfigure}{0.08\textwidth}
            \includegraphics[width=\linewidth]{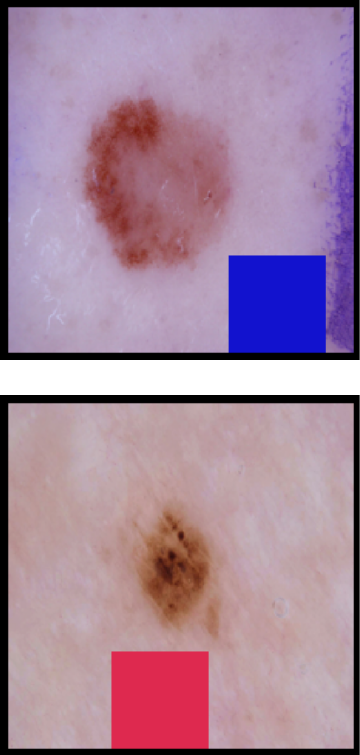}
            \end{subfigure}
            \begin{subfigure}{0.08\textwidth}
            \includegraphics[width=\linewidth]{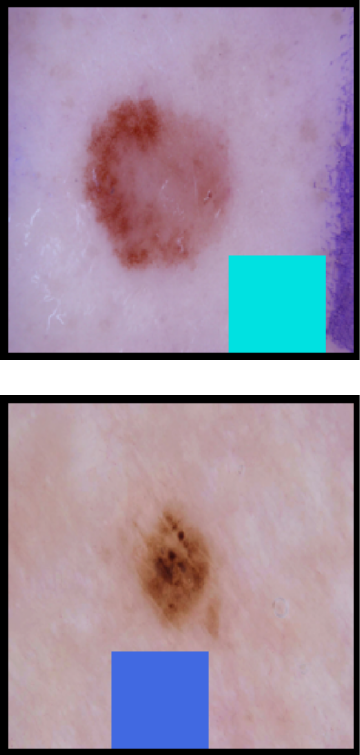}
            \end{subfigure}
            \begin{subfigure}{0.08\textwidth}
            \includegraphics[width=\linewidth]{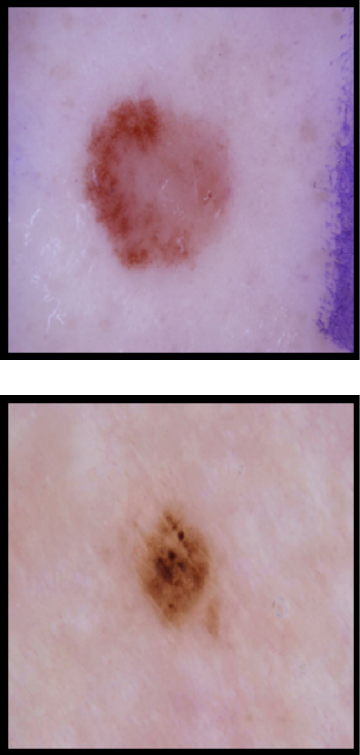}
            \end{subfigure}
            \begin{subfigure}{0.08\textwidth}
            \includegraphics[width=\linewidth]{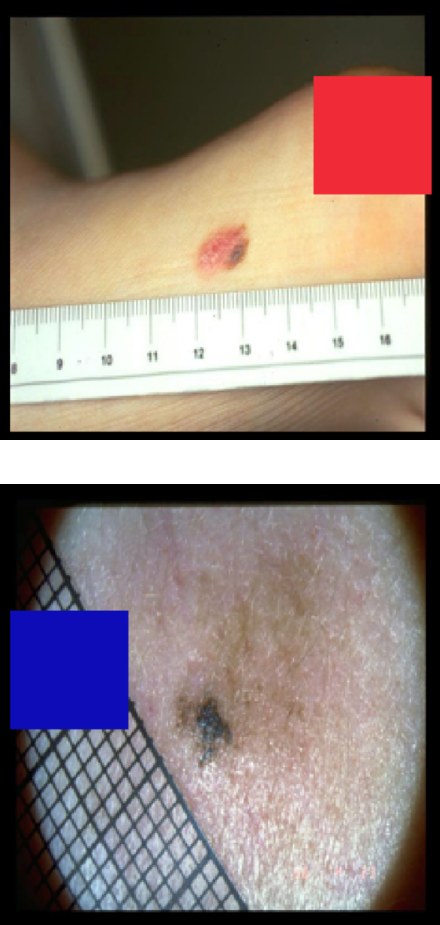}
            \end{subfigure}
            \begin{subfigure}{0.08\textwidth}
            \includegraphics[width=\linewidth]{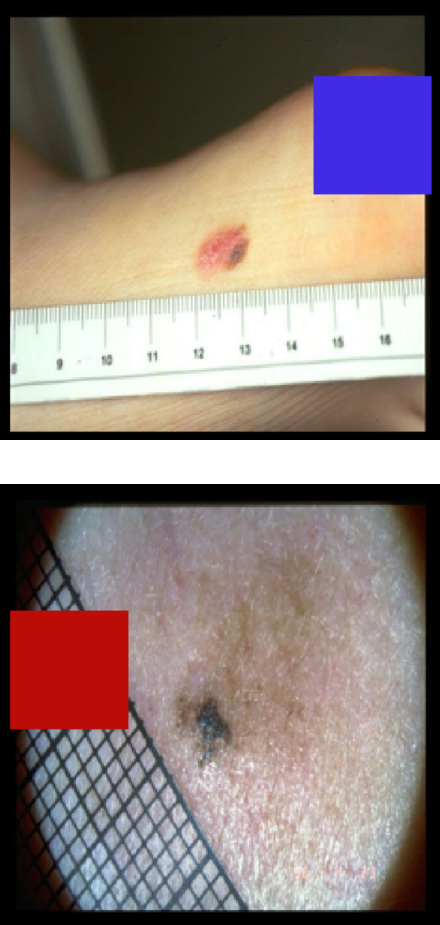}
            \end{subfigure}
            \begin{subfigure}{0.08\textwidth}
            \includegraphics[width=\linewidth]{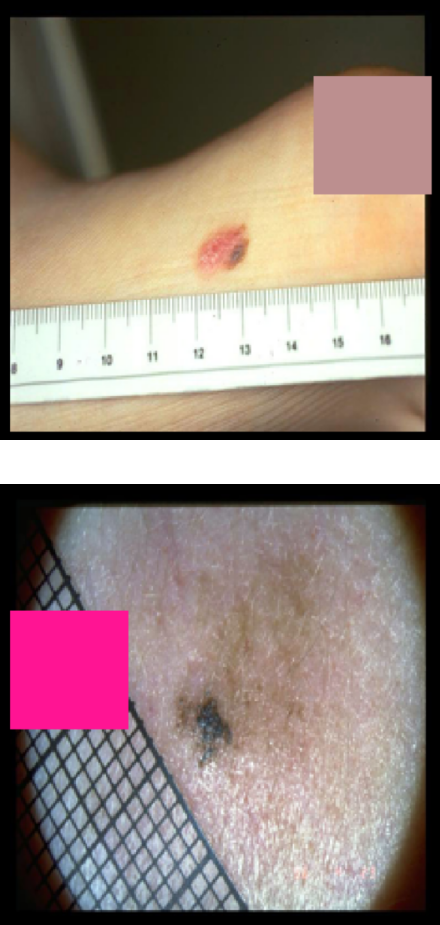}
            \end{subfigure}
            \begin{subfigure}{0.08\textwidth}
            \includegraphics[width=\linewidth]{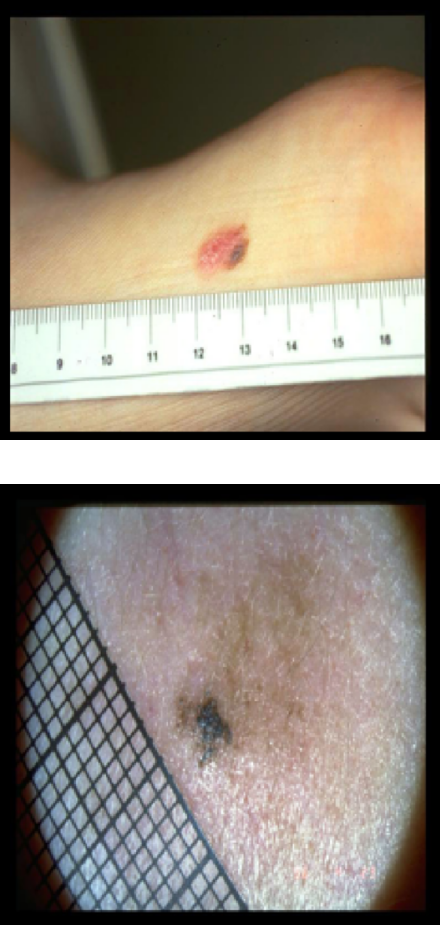}
            \end{subfigure}

        \vspace{0.1cm}
        \textbf{\footnotesize{Real correlation and diversity shifts}}\\
        
        \begin{subfigure}{0.08\textwidth}
            \includegraphics[width=\linewidth]{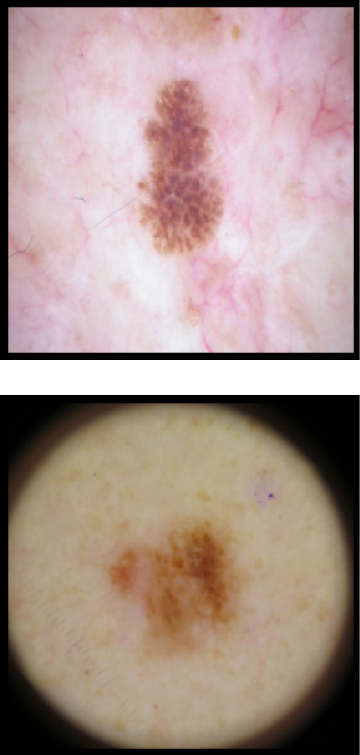}
            \caption{}
        \end{subfigure}
        \begin{subfigure}{0.08\textwidth}
            \includegraphics[width=\linewidth]{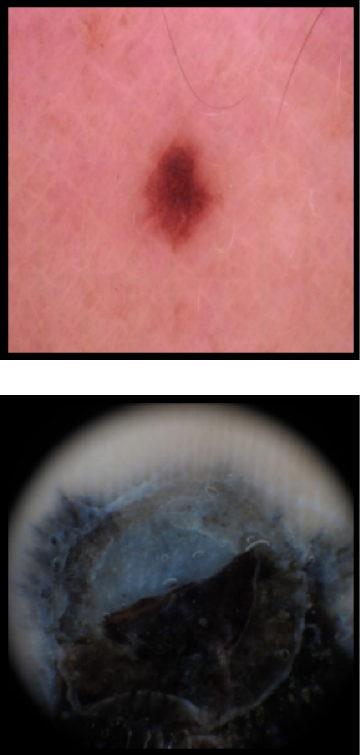}
            \caption{}
        \end{subfigure}
        \begin{subfigure}{0.08\textwidth}
            \includegraphics[width=\linewidth]{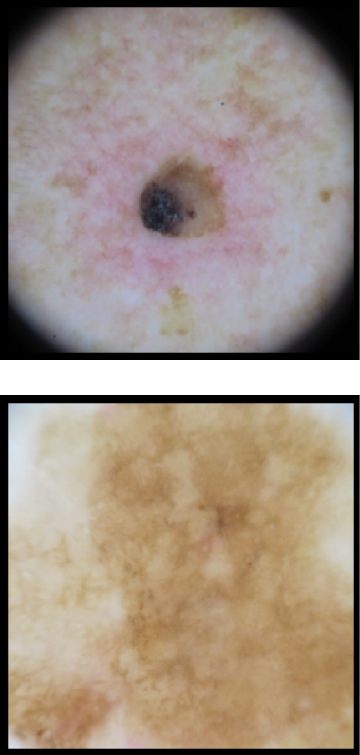}
            \caption{}
        \end{subfigure}
        \begin{subfigure}{0.08\textwidth}
           \includegraphics[width=\linewidth]{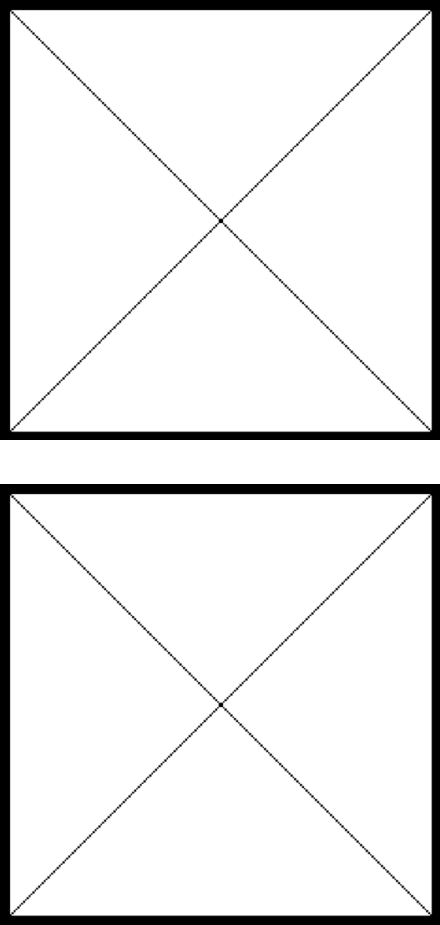}
            \caption{}
        \end{subfigure}
        \begin{subfigure}{0.08\textwidth}
            \includegraphics[width=\linewidth]{cross.pdf}
            \caption{}
        \end{subfigure}
        \begin{subfigure}{0.08\textwidth}
            \includegraphics[width=\linewidth]{cross.pdf}
            \caption{}
        \end{subfigure}
        \begin{subfigure}{0.08\textwidth}
            \includegraphics[width=\linewidth]{cross.pdf}
            \caption{}
        \end{subfigure}
        \begin{subfigure}{0.08\textwidth}
            \includegraphics[width=\linewidth]{squared_skin_tdn.pdf}
            \caption{}
        \end{subfigure}
        \begin{subfigure}{0.08\textwidth}
            \includegraphics[width=\linewidth]{cross.pdf}
            \caption{}
        \end{subfigure}
        \end{subfigure}
    \end{minipage}%
    \caption{Example of our training and test sets for each of proposed experiment. The complexity of our datasets progressively grows from the top to the bottom, both in terms of the available correlation shifts, which start with colored squares providing spurious features and end with image acquisition artifacts from a real-world problem; and also in terms of the available robust features, as it ranges from simpler general purpose object classification to skin lesion analysis.
    On training \textbf{(a)}, when possible, we control the intensity of bias by manipulating the correlations between colors and labels. Fig.~\textbf{(b)} to \textbf{(e)} show our test samples without diversity shift, and test scenarios \textit{same-same}, \textit{same-diff}, \textit{diff}, and \textit{no shortcuts}. The same scenario sequence repeats from \textbf{(f)} to \textbf{(i)} for diversity shift test sets. For the real correlation and diversity shifts, we lack correspondences for the \textit{i.i.d diff} test set and only have diff for diversity-shifted sets.}
    \label{fig:dataset_squares}
\end{figure*}

\subsection{Data modification}

A comprehensive distribution shift analysis requires controlling the levels of spurious correlations during training and evaluating models on carefully designed test sets that measure both the exploitation of shortcuts (correlation shift effect) and generalization capabilities for diversity-shifted data. We propose two synthetic modifications to enable this analysis.

\vspace{0.1cm}
\noindent{\textbf{\textit{Synthetic correlation shift.}}} \label{corrshift}
To control correlation shift, we introduce shortcuts as colored squares in \textbf{every} training sample. During training, we assign a square color for each class and control the bias intensity through what we term ``training biases'', representing the percentage of the training set spuriously correlated with a given color. For instance, in a binary classification task between classes A and B, a training bias of 70 signifies a dataset where $70\%$ of class A samples display a blue square and $30\%$ a red one, while for class B, $70\%$ show a red square and $30\%$ a blue one. Similarly, an unbiased set has a $50$ training bias (e.g., $50\%$ of samples from both classes exhibit red squares and the remaining blue ones), and a $100$ training bias set has both spurious and invariant features being fully predictive of the~label.

To make the shortcut more challenging to learn, we randomly place colored squares on one of the image borders (avoiding the occlusion of relevant information) and apply noise to the biasing colors, resulting in slightly varying color hues across images. We maintain a constant square size of approximately 8\% of the image, as our experiments have shown that size is not a significant factor during training.

For testing, inspired by Ahmed et al.~\cite{ahmed2020systematic}, we introduce different scenarios: \textbf{1.} same colors and same coloring scheme from training \textit{(same-same)}, \textbf{2.} same colors but different coloring scheme \textit{(same-diff)}, \textbf{3.} \textit{no shortcuts}, and \textbf{4.} different colors from training \textit{(diff)}. Each of these four schemes has a variant with (Fig.~\ref{fig:dataset_squares} (f) to (g)) and without diversity shift (Fig.~\ref{fig:dataset_squares}~(b)~to~(e)).

\vspace{0.1cm}
\noindent{\textbf{\textit{Synthetic diversity shift.}}}
To manage diversity shift, we exploit distribution differences across subclasses. For example, consider a \textit{flowers} superclass consisting of subclasses \textit{orchids, poppies, roses, sunflowers}, and \textit{tulips}. We can use this hierarchy to create our train and test sets. One possible division is to select \textit{orchids, poppies, and roses} for our training and in-distribution (i.i.d.) test, while \textit{sunflowers and tulips} are used for test sets with diversity~shift.

By following a given hierarchy (e.g., WordNet~\cite{miller1998wordnet}), we first organize the data into superclasses composed of at least five subclasses, using three for training and the remaining two for testing. We design our classification problems to be binary and use superclass divisions to extract multiple tasks from large datasets. For fair comparisons, we restrict our test sets to the same size for a given experiment.

\subsection{Experimental Design}

In this section, we detail our three distinct binary classification experiments, each with increasing complexity. We adapt existing datasets to exhibit controllable correlation and diversity shifts using the previously described strategies when necessary. We believe our binary classification problems are representative of problems in the wild, and expanding this analysis to the multi-class case is left for future work.

\vspace{0.1cm}
\noindent{\textbf{\textit{Synthetic correlation and diversity shifts.}}}
We partition the ImageNet dataset~\cite{ILSVRC15} into $9$ random binary classification problems\footnote{The full set of problems studied are: mammals \textit{vs.} domestic dog, construction \textit{vs.} insect, automotive vehicle \textit{vs.} green goods, implement \textit{vs.} garment, aliment \textit{vs.} transport, covering \textit{vs.} equipment, mammals \textit{vs.} covering, transport \textit{vs.} vessel, aliment \textit{vs.} construction.}, utilizing the WordNet hierarchy~\cite{miller1998wordnet} to define superclasses. This approach ensures varying levels of task difficulty and feature diversity. We introduce correlation shifts by adding colored squares, and we achieve diversity shifts by exploiting the target classification's subclasses (see Section \ref{corrshift}).
The training sets consist of $2400$ images, while validation sets contain $600$ images, and test sets comprise $300$ images each.

\vspace{0.1cm}
\noindent{\textbf{\textit{Synthetic correlation shift and real diversity shift.}}}
For skin cancer classification, we utilize the HAM10000 dataset~\cite{tschandl2018ham10000} in a melanoma vs. benign binary classification task as training and test data without diversity shift. We employ the BCN20000~\cite{combalia2019bcn20000} and Derm7pt-Clinical~\cite{Kawahara2018-7pt} datasets as diversity-shifted test sets\footnote{We designate melanoma as our only malignant target, removing all basal cell carcinoma samples from the datasets.}. These test sets represent different degrees of diversity shifts. BCN20000, which is closer to the training distribution, contains dermoscopic images\footnote{Dermoscopic images are captured using a specialized image acquisition device called a dermatoscope, which reduces light interference and enables physicians to analyze dermoscopic features.} from different hospitals, while the Derm7pt-Clinical dataset, an older collection created for educational purposes, features images captured with conventional cameras. All sets exhibit a natural class imbalance, with benign images being more prevalent than malignant ones, and the imbalance ratio varies across datasets. We introduce correlation shifts using colored squares (see Section~\ref{corrshift}). The HAM10000 dataset is divided into $6128$ training samples and $1426$ validation samples; BCN20000 contains $8201$ samples, and Derm7pt-Clinical comprises $839$ samples.

\vspace{0.1cm}
\noindent{\textbf{\textit{Real correlation and diversity shifts.}}} 
To ensure our findings are not limited to the synthetic settings produced by our dataset adaptation procedures, we examine the results on an unmodified dataset for skin lesion analysis. We map the results of Bissoto et al.~\cite{bissoto2022} onto our framework, which is only possible due to the availability of publicly accessible out-of-distribution test sets~\cite{Kawahara2018-7pt, pacheco2020pad} and the data organization in increasing levels of bias~\cite{bissoto2020debiasing}. Bissoto et al.~\cite{bissoto2020debiasing} annotated the ISIC 2019 dataset~\cite{isic2019data}, a conventional skin lesion analysis dataset containing $25,331$ images, with respect to the seven existing artifacts\footnote{Dark corners, rulers, hair, ink markings, gel bubbles, gel borders, and patches.} introduced during image acquisition. 
This approach enables the creation of what the authors refer to as ``trap sets''. In these sets, both \textit{trap train} and \textit{trap validation} contain amplified spurious correlations, while \textit{trap test} exhibits correlation shifts. Trap sets are obtained through an optimization process that maximizes the separation between training and test sets concerning artifact presence. Intuitively, this process attempts to allocate, for example, most malignant samples with dark corner artifacts to the training set and most benign samples with dark corners to the test set. A model that learns to associate dark corners with lesion malignancy will perform poorly on the test set. Different bias intensities can then be obtained by interpolating between random and found trap train-test division.

Like our analysis, trap sets allow for controlling training biases and evaluating \textit{i.i.d. same-same} and \textit{same-diff} performances using \textit{trap validation} and \textit{trap test} sets, respectively. However, since trap sets amplify and control bias through data partitioning, images from train and test sets vary across training biases, potentially causing uncertainties when adapting this procedure to our evaluation protocol. Nonetheless, we recognize that such uncertainties are inherent to real-world scenarios and justify our previous study cases involving synthetic correlation~shifts.

Lastly, Bissoto et al.~\cite{bissoto2022} assess their method on out-of-distribution sets derm7pt-clinical~\cite{Kawahara2018-7pt} and padufes~\cite{pacheco2020pad}, introducing shifts concerning different populations, diagnoses, and image modalities. As observed in their study \cite{bissoto2022}, biases across out-of-distribution datasets for skin lesion analysis vary significantly, positioning these sets as \textit{diversity-shifted diff} test sets in our evaluation.

In this experiment, we lack correspondences for the \textit{i.i.d. diff} test set and only have \textit{diff} for diversity-shifted sets (see blank square examples in Fig.~\ref{fig:dataset_squares}). To fill these missing cases, we would need to collect samples from the same patients with and without artifacts (for \textit{no shortcuts}) and change the source hospital for diversity shift while maintaining or modifying previous image acquisition protocols (for diversity-shifted \textit{same-same} and \textit{same-diff}). This process does not occur naturally and would require collaboration between physicians and machine learning experts, making it both expensive and labor-intensive.

\subsection{Experimental details}

We design our experiments to cover multiple datasets, with different types and intensities of correlation and diversity shifts. In our scenarios, we cover $8$ training biases from $52$ to $80$, in increments of $4$.  Unlike previous research, we consider low levels of biases.

We also include in our study common factors that may influence the ability of models to generalize, such as model architecture, and pretraining on a bigger dataset (e.g., ImageNet). 
 For the experiments with synthetic correlation shifts, we employ a ResNet-18 model \cite{he2016deep}. 
 This is a common choice in the domain generalization literature~\cite{ye2021ood}, and its fast training and inference time allow us to run 10 replicas for improved statistical significance. In the real-world skin lesion case, a ResNet-50 model was used~\cite{bissoto2022}, showing our findings are also present in deeper models. 
 We fine-tune ImageNet-pretrained models for skin lesion contexts, and train models from scratch for ImageNet.
To ensure a challenging environment for our experiments closer to a real-world scenario, we select all hyperparameters on a validation set from the same distribution of training, assuring our models never had privileged access to test information or to data distributions where biases are absent or balanced.

\begin{figure*}[ht]
    \centering
    \includegraphics[width=0.95\linewidth]{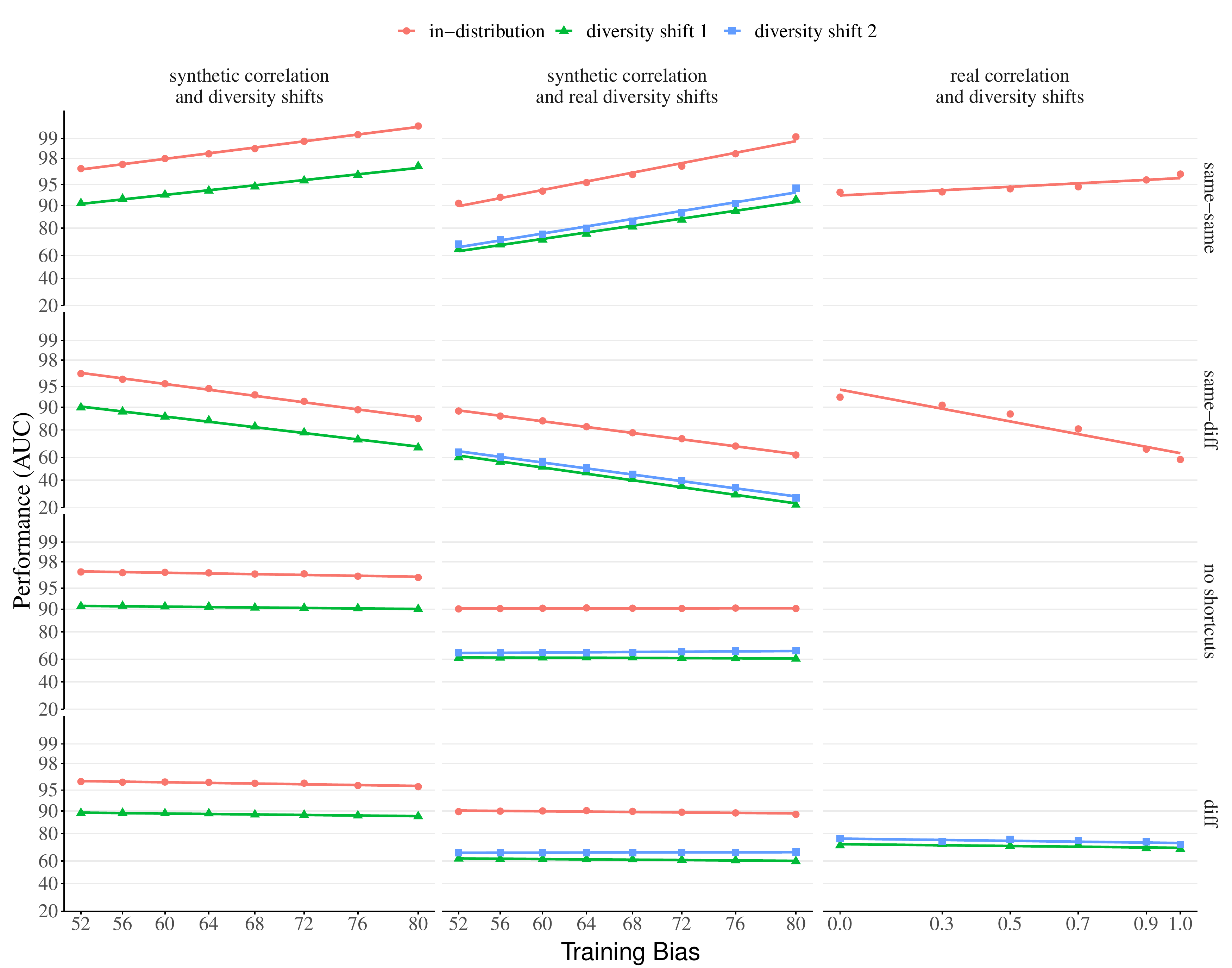}

    \caption{
    Each column represent one of our experiments, organized in increasing order of complexity. Each row display each of our test sets. Each line hue represent in-distribution or diversity-shift sets\protect\footnotemark. 
    Lines represent the fitted linear model for each test set. Each point represent the average of $10$ runs. Our findings across different experiments and scenarios show that reliance on biases occurs even in very low-biased scenarios (low training bias on \textit{same-same} and \textit{same-diff} curves), and, despite biased training sets, a model can still yield robust and accurate predictions if the shortcuts are absent or different on test images (\textit{no shortcuts} and \textit{diff} curves). For real correlation and diversity shifts, we keep the training bias scale the same as the source work~\cite{bissoto2020debiasing}.
     }
    \label{fig:realworld}
\end{figure*}

\section{Results and Analysis}
    
In Fig.~\ref{fig:realworld}, we show a grid of our results. Each column represent a experiment (i.e., \textit{synthetic correlation and diversity shifts}, \textit{synthetic correlation shift, and real diversity shift}, and \textit{real correlation and diversity shifts}), and each row represent a different type of test set (i.e., \textit{same-same}, \textit{same-diff}, \textit{no shortcuts}, \textit{diff}). The line hues identify in-distribution and diversity-shifted test~sets.

We also evaluated training models
using balanced accuracy instead of AUC (area under the curve) to measure performance, and adding groupDRO \cite{sagawa2019distributionally} as a training algorithm. 
Considering all these scenarios and configurations, our findings remained. 
Next, we discuss each finding separately.

\vspace{0.1cm}
\noindent{\textbf{\textit{Low-biases are the most problematic.}}}
Increasing training bias directly affect the performances for \textit{same-same} and \textit{same-diff} test sets. As expected, performances for the former increase, and decrease for the latter, showing models' reliance on the introduced spurious feature. 
However, it is very concerning that performances increase and decrease linearly (in the \textit{logit} scale): bias reliance can dominate the prediction if the training bias is strong enough; and more importantly, it affects solutions even in scenarios with mild biases (Fig.~\ref{fig:realworld} on training biases from $52$ to $60$ for synthetic correlation shifts, and from $0$ to $0.5$ for real correlation shifts).

This ability of deep neural models to learn and memorize data is not new~\cite{zhang2021understanding}. This principle was shown to be responsible for pretraining success in improving generalization, as larger datasets used for pretraining often include at least a few counterfactual examples to previously existing biases \cite{tu2020empirical}.
Our experiments show that this extraordinary ability of models to incorporate infrequent patterns can act as a double-edged sword, as models exploit even weak spurious correlations. 
Scaling data to balance datasets for all possible confounders is unfeasible for non-synthetic data. For example, despite the colossal amount of data and parameters, GPT-3 still reproduces biases found in its training data even in the presence of counterfactual examples~\cite{brown2020language}. 
Moreover, scaling models and data is not always an option. In critical contexts (e.g., medical), scaling the size of datasets is often impractical due to the costs of acquiring good quality annotated data. For attenuating bias, providing additional annotations that empower domain generalization methods must become the new standard \cite{daneshjou2022checklist}.

\vspace{0.1cm}
\noindent{\textbf{\textit{Models learn robust features even in high-bias scenarios.}}}
When the test set does not present precisely the same shortcuts as training, the performance remains stable even when the training set is heavily biased. This behavior is verified by looking at the \textit{diff} and \textit{no shortcuts} rows in Fig.~\ref{fig:realworld}: For synthetic correlation shifts, both removing the biasing squares or coloring them in colors unseen during training achieves this stabilizing effect. For real correlation shifts, we verify the same effect when evaluating on out-of-distribution sets.
Contrary to what was~previously thought~\cite{pezeshki2020gradient}, the training models did not abdicate to learn correct features alongside the spurious ones, being able to classify unbiased samples in biased models (almost) as well as in unbiased ones. 
Previous work identified the presence of unbiased subnetworks in biased models \cite{zhang2021can}, but accessing them required heavy instrumentation in the model. 
We show that as long as spurious patterns are not available in test images, or are different from the ones learned during training, models yield robust predictions.

\footnotetext{In \textit{synthetic correlation and diversity shifts}, we have three ImageNet subclasses as \textit{in-distribution}, and two different ones as \textit{diversity shift 1}.
In \textit{synthetic correlation and real diversity shifts} we have HAM10000 as \textit{in-distribution}, and derm7pt-clinical, and BCN20000 as\textit{ diversity shifts 1 and 2}, respectively. 
For \textit{real correlation and diversity shifts} we have ISIC2019-trap as \textit{in-distribution}, and derm7pt-clinical and padufes as \textit{diversity shift 1 and 2}, respectively.}

\vspace{0.1cm}
\noindent{\textbf{\textit{Diversity shift attenuates correlation shift.}}}
The performance curves are less steep when diversity shift is present in the test set.
For quantifying reliance over shortcuts, we evaluate the angular coefficient of the regression line for ``performance $\sim$ training bias''. As training bias increases, the angular coefficient captures the direction (positive or negative value) and intensity of performances' variation. High absolute values indicate that model's performance is highly affected by the training bias, while close-to-zero coefficients indicate robustness to the shortcut introduced during training.
With this measurement, we compare the performance reached in different test types and contrast sets regarding the presence of diversity shift.

\begin{table}[th!]
\caption{Angular coefficients of the linear regression for ``performances $\sim$ training biases'' for test sets with and without diversity shift on the \textit{synthetic correlation and diversity shifts} context with ImageNet. Surprisingly, diversity shift tests present closer-to-zero angular coefficients, indicating lower reliance on shortcuts (highlighted in bold in the table).}
\footnotesize
\centering
\label{tab:coefs}
\begin{tabular}{lcccc}
\toprule
                & same-same & same-diff & no shortcuts & diff \\
\midrule
in-distribution & 1.17     & 1.22      & 0.13         & 0.14 \\
diversity-shift & \textbf{0.98}      & \textbf{1.10}      & \textbf{0.10}         & \textbf{0.08} \\
\bottomrule
\end{tabular}
\end{table}

Focusing on the \textit{synthetic correlation and diversity shifts} context with ImageNet, our results show diversity-shifted test sets present angular coefficients closer to zero than their non-shifted counterparts (Table~\ref{tab:coefs}).
This non-intuitive effect suggests that diversity shift has an attenuating effect on correlation shift. At inference time, models use the learned weights to extract and compress learned patterns into features used for classification. However, when facing novel, previously unseen patterns during inference (as in a diversity shift scenario), we expected models to rely upon more --- not less --- on the available shortcuts learned during training \cite{geirhos2020shortcut}.

\section{Conclusion}

In this paper, we provide an extensive and comprehensive analysis of the effects of diversity and correlation shifts on deep learning models, validated in synthetic and real-world datasets with different levels of complexity. 
Our findings challenge the beliefs of current distribution shift research, pointing paths towards more realistic and integrated research where biases co-exist.
Our main finding is that low-biases can have a significant effect on deep learning models. 
Even extremely low biases, where the probability of presenting the spurious feature is slightly higher than chance, are sufficient to poison models. This is particularly problematic considering the ability of models to extract non-semantic features from data~\cite{ilyas2019adversarial}. 

Despite this significant influence of biases during training, we also found that models are able to learn robust features in both low and high bias scenarios. 
We think exploiting this can lead to more robust models. 
A possible way forward is using test-time debiasing~\cite{niu2023towards} to filter robust features from the spurious ones, or to align source and target distributions. 
Better annotations~\cite{daneshjou2022checklist} and methods for discovery of undesired bias~\cite{creager2021environment} can boost existing domain generalization methods to increase robustness.
Finally, causal representation learning \cite{scholkopf2021toward} investigate solutions to encourage a causal structure in the latent space, enabling more transparent and debiased solutions.

\section*{Acknowledgments}
A. Bissoto is funded by FAPESP (2019/19619-7 and 2022/09606-8). 
C. Barata is funded by the FCT projects LARSyS (UID/50009/2020) and CEECIND/00326/2017.
E. Valle is partially funded by CNPq 315168/2020-0. 
S.~Avila is partially funded by CNPq 315231/2020-3, FAPESP 2013/08293-7, 2020/09838-0, Google LARA 2021 and Google AIR 2022. 


\bibliographystyle{elsarticle-num} 




\end{document}